\documentclass[letterpaper]{article}
\usepackage{aaai2026}
\usepackage{times}
\usepackage{helvet}
\usepackage{courier}
\usepackage[hyphens]{url}
\usepackage{graphicx}
\usepackage{natbib}
\usepackage{caption}
\usepackage{bbold}
\usepackage{amsmath}
\usepackage{algorithm}
\usepackage{algorithmic}

\title{How to Marginalize in Causal Structure Learning?}

\author {
    William Zhao,
    Guy Van den Broeck,
    Benjie Wang
}
\affiliations {
    University of California, Los Angeles \\
    wizhao@ucla.edu, guyvdb@cs.ucla.edu, benjiewang@cs.ucla.edu
}

\begin{document}
\maketitle
\begin{abstract} 
    Bayesian networks (BNs) are a widely used class of probabilistic graphical models employed in numerous application domains. However, inferring the network’s graphical structure from data remains challenging. Bayesian structure learners approach this problem by inferring a posterior distribution over the possible directed acyclic graphs underlying the BN. The inference process often requires marginalizing over probability distributions, which is typically done using dynamic programming methods that restrict the set of possible parents for each node. Instead, we present a novel method that utilizes tractable probabilistic circuits to circumvent this restriction. This method utilizes a new learning routine that trains these circuits on both the original distribution and marginal queries. The architecture of probabilistic circuits then inherently allows for fast and exact marginalization on the learned distribution. We then show empirically that utilizing our method to answer marginals allows Bayesian structure learners to improve their performance compared to current methods.
\end{abstract}

\section{Introduction}
Bayesian networks (BNs) are a popular choice for representing probabilistic relationships between variables. Given that they have widespread benefits in interpretability and transparency compared to other more black box approaches, they have been employed in a wide variety of applications, ranging from healthcare modeling to industrial fault diagnosis. \cite{kyrimia2021bayesiannetworkshealthcare, cai2017bayesianfaultdiagnosis}. In a BN, variables are represented as nodes in a graph, with directed edges representing conditional dependencies between two variables. Collectively, these edges form a directed acyclic graph (DAG) across all variables. 

Structure learning is the task of discovering the DAG structure of a BN given data. However, structure learning poses a difficult computational challenge, since the number of DAGs grows super-exponentially with respect to the number of variables $n$ \cite{kitson2023structurelearningsurvey}.
To account for uncertainty among the possible DAG structures, \emph{Bayesian} structure learning methods aim to infer the posterior distribution of possible graphs $G$. These structure learning methods are based on Bayesian scoring functions, which output (possibly unnormalized) posterior probabilities of graphs given data. 

Within Bayesian structure learning, marginalizing these scores over all possible parent sets of a node serves as a fundamental subtask. Prior methods have relied on exact calculation with a dynamic programming table. However, since this dynamic programming computation requires exponential time in the number of candidate parents, these methods artificially limit the number of candidate parents to a set of the most likely parents. This method thus lacks support over all possible parent sets, which limits the performance of the structure learner. 

Probabilistic circuits are a fitting architecture for replacing dynamic programming for this subtask. Probabilistic circuits are a type of tractable probabilistic model that, when satisfying certain structural properties, allow for efficient computation of any possible marginal and conditional query \cite{choi2020probcircuits}. Thus, we learn circuits that approximate the distribution represented by the Bayesian scores and then use them as a surrogate to answer the queries required by the structure learner.

We present an approach that trains a probabilistic circuit per node to answer the marginal and sampling queries necessary for Bayesian approaches to structure learning. This method utilizes a new learning algorithm that employs as ground truth both unmarginalized probability masses and marginalized probability masses calculated via a more limited exact brute force approach. This method improves upon prior approaches that require limits on the number of potential parents for each node. On a specific Bayesian structure learning method, TRUST \citep{wang2022trust}, we demonstrate empirical improvements to results utilizing our method. 

\section{Related Work}
Given the superexponential search space over all possible structures, one common strategy for structure learners is to simplify the search space into one of topological orders. This approach has been used in a wide variety of structure learning approaches, including ones that utilize exact search \cite{teyssier2005orderingbasedsearch}, MCMC approximation \cite{friedman2003beingbayesianaboutnetwork, kuipers2017partitionmcmc, viinikka2020scalablebayesianlearning}, and gradient-based approaches \cite{wang2022trust, cundy2021bcdnets, annadani2023bayesdag, toth2025effectivebayesiancausalinference}. 

An especially attractive property of a topological ordering is that by fixing an ordering, we fix the set of possible parents for each node. However, even this limitation is not enough for many methods, which instead place further restrictions on a node's parent sets for computational reasons. \citet{kuipers2017partitionmcmc} and \citet{toth2025effectivebayesiancausalinference}, for example, forcibly restrict the size of any possible parent sets. Meanwhile, \citet{viinikka2020scalablebayesianlearning} and \citet{wang2022trust} instead produce a limited set of possible candidate candidates for each node: for any node, the set of parents must be a subset of that candidate set. Unfortunately, these restrictions limit the search space of the method and thus the structures that may actually be produced by the structure learner.

There are many different approaches to structure learning of Bayesian networks. Of particular interest are Bayesian approaches to structure learning, which infer a posterior distribution for the structure of a Bayesian network given data. Learning this distribution allows for a variety of useful queries, including Bayesian model averaging of causal effects \cite{viinikka2020scalablebayesianlearning}. Exact methods have not scaled above roughly $d=20$ \cite{koivisto2006exactbayesian}, which has driven work towards approximate methods. Such works notably include \citet{friedman2003beingbayesianaboutnetwork}, which utilized MCMC sampling over the distribution of orders. TRUST, an approximate Bayesian structure learning method introduced in \citet{wang2022trust}, is utilized as a test case for our method. We chose TRUST because it already employs probabilistic circuits for representing the posterior distribution over orderings. Thus, by incorporating the method we present as well, we have developed a structure learner entirely reliant on probabilistic circuits.

Probabilistic circuits were introduced in \citet{choi2020probcircuits} as a class of probabilistic models that use computational graphs to encode probability distributions. Compared to other models, probabilistic circuits allow for efficient inference, including exact marginalization, on the distributions they learn. The tractability of probabilistic circuits has been exploited for tasks such as constrained language generation \citep{zhang2024adaptable,yidou2025trace,ahmed2025semantic}, neuro-symbolic AI \citep{ahmed2022semantic,loconte2023turn,wang2024compositional,karanam2025unified}, and causal reasoning \citep{wang2021provable,wang2023compositional,busch2024psi}. Probabilistic circuits are frequently learned from distributions, and making this learning more efficient and more scalable has been the subject of much work \cite{peharz2020einsumnetwork, loconte2024subtractive, wang2025relationship, zhang2025monarchmatrices}. Other architectures that focus on answering marginalization queries include NADEs \cite{uria2014deepdensityestimator}, AO-ARMs \cite{shih2022training} and MAMs \cite{liu2024generativemarginalizationmodels}. However, these methods only approximate these queries on their learned distributions and do not provide exact inference, unlike probabilistic circuits.

\section{Background}

\subsection{Marginalization}
Let $p(\mathbf{X})$ be an unnormalized probability distribution, where $\mathbf{X}$ is a vector of $n$ binary variables $\{x_0, x_1, \dotsc,  x_n\}$. Given that most Bayesian scores represent unnormalized posteriors, we do not assume that $\sum p(\mathbf{X}) = 1$.

Let $\mathbf{X}_V$ and $\mathbf{X}_{V_C}$ be subvectors of $\mathbf{X}$ corresponding to indices $V \subseteq \{1, 2, 3, \dotsc, n \}$ and the complement $V_C = \{1, 2, 3, \dotsc, n \} \setminus V$. A marginal distribution over $\mathbf{X}_V$
can be defined by summing over all possible values of $\mathbf{X}_{V_C}$: \[p(\mathbf{X}_V) = \sum_{\mathbf{X}_{V_C}} p(\mathbf{X}_V, \mathbf{X}_{V_C}) \]

\subsection{Probabilistic Circuits}
Probabilistic circuits (PCs) are a class of tractable probabilistic models defined by a computational graph with a rooted DAG structure, where each node represents a computational unit \cite{choi2020probcircuits}. Given a circuit over a set of variables $\mathbf{X} = \{x_1, x_2, \cdots , x_n\}$, each node represents a distribution over a subset of $\mathbf{X}$. This subset of variables is considered to be the scope of the node $sc(v)$. Within a circuit, there are several different types of nodes:

\begin{itemize}
    \item Leaf nodes represent input probability distributions over a single variable; in our case of binary variables, these are Bernoulli distributions with some flip probability $p$.
    \item Product nodes take the product of the values of their child nodes. Thus, the value of a product node $v_i = \prod_{j \in Ch(i)} v_j$, where $Ch(i)$ are all children of node $i$.
    \item  Sum nodes take the weighted sum of the values of their child nodes. Thus, the value of a sum node is $v_i = \sum_{j \in Ch(i)} w_jv_j$, where $Ch(i)$ are all children of node $i$ and $w_j$ correspond to the weight of node $j$ in the sum. In many circumstances, PCs enforce normalization, i.e., $\sum w_j = 1$. However, we do not enforce this constraint in this paper.
\end{itemize}

The probability distribution of the PC is then defined by the distribution represented by the root node. In particular, evaluating $p(\mathbf{X})$ simply requires evaluating the leaf probabilities for that assignment and then computing the sums and products in a forward pass.

Circuits with certain structural properties prove to be significantly more tractable. \textit{Smoothness} and \textit{decomposability} in particular are highlighted here \cite{poon2011spn}:
\begin{itemize}
\item A PC holds the property of \textit{smoothness} if for all sum nodes $S$ and for all children $C_1, C_2$ of $S$, $sc(C_1) = sc(C_2)$. In a smooth PC, children of the same sum node have the same scope. 
\item A PC holds the property of  \textit{decomposability} if for all product nodes $P$ and for all children $C_1, C_2$ of $P$, $sc(C_1) \cap  sc(C_2) = \emptyset$. In a decomposable PC, children of the same product node have pairwise disjoint scopes.
\end{itemize}

If a PC is both decomposable and smooth, PCs can exactly marginalize over their represented distribution in time proportional to the size of the PC. Thus, PCs with these properties are particularly computationally efficient and tractable compared to alternative architectures.

Moreover, PCs can be viewed as a neural network with sum nodes providing linear transformations on their input values and leaf and product nodes contributing nonlinearity \cite{choi2020probcircuits}. Thus, it is possible to follow the conventional architecture of a neural network of linear layers followed by nonlinear activations by alternating sum layers with product layers. Accordingly, PCs can be trained to learn a distribution using standard gradient descent techniques utilizing backpropagation.

\subsection{Bayesian Structure Learning}
A Bayesian network $\mathcal{N}$ over a set of variables $\mathbf{X} = \{ X_1, X_2, \hdots, X_n\}$ can be defined as a tuple $(G, \Theta)$. $G$ is a directed acyclic graph, where edges represent conditional independence between variables in the network. $\Theta$ represents a set of parameters that define the conditional probability distributions between variables. Let the parents of a variable $X_i$ in the graph $G$ be denoted $pa_G(X_i)$. Since the conditional probability distribution for a variable $X_i$ is then $p(X_i \vert pa_G(X_i), \Theta_i)$, we obtain the joint probability distribution for the BN $\mathcal{N}$:

\[p(\mathbf{X} \vert G, \Theta) = \prod_{i=1}^{n}  p(X_i \vert pa_G(X_i), \Theta_i)\]

\textit{Structure learning} is the process of learning the DAG $G$ of a BN that generates specific data $D$. When approaching structure learning with Bayesian methods, we define a structural prior $p_{pr}(G)$ and a likelihood $p_L(D|G)$. \textit{Modularity}, or that priors and likelihoods can be written as products over nodes, is a common assumption \cite{eggeling2019structurepriors}. The posterior can thus be written as:
\[p(G\vert D) \propto  \prod_{i=1}^{n} \mathbb{1}_{\textit{DAG}(G)} \, p_{pr} (pa_G(X_i)) \, p_{L} (D \vert pa_G(X_i))\]
where $\mathbb{1}_{\textit{DAG}(G)}$ is an indicator variable that represents whether $G$ is acyclic. This assumption simplifies the problem significantly, as it allows every node's set of parents independently to be considered independently from each other given the acylicity constraint (for instance, once an ordering is imposed).

Bayesian scores $\mathcal{B}$ are a class of scores whose values are proportional to the posterior probabilities of structures \cite{kitson2023structurelearningsurvey}. Each specific Bayesian score assumes a prior distribution over network parameters. Different Bayesian scores are specialized for different types of Bayesian networks because they require different parameter priors. For example, BGe scores are used for BNs with linear Gaussian mechanisms, and BDeu scores are used for BNs with discrete mechanisms. These scores are generally unnormalized, and for numerical stability reasons, these scores are given in the log domain. Thus, given a directed acyclic graph, they compute:
\[\mathcal{B}(G) = \sum_{i=1}^{n} \log p_{pr}  (pa_G(X_i)) + \log p_{L} (D \vert pa_G(X_i))\]
Note that including the structural prior $p_{pr}$ in $\mathcal{B}$(G) is optional and that in some literature the specific Bayesian score refers to the likelihood $p_L$ only \cite{eggeling2019structurepriors}. We include this term as in the work of \citet{kitson2023structurelearningsurvey} to simplify our notation.

Since we assumed structural modularity, we can then decompose the score of a
graph $\mathcal{B}(G)$ as a sum of scores $\mathcal{B}_{X_i}(pa_{G}(X_i))$ associated with a node and its parent set, where:
\[\mathcal{B}_{X_i}(pa_{G}(X_i)) = \log p_{pr}  (pa_G(X_i)) + \log p_{L} (D \vert pa_G(X_i))\]
For simplicity, we can also represent the input $pa_{G}(X_i)$ as a vector of indicator variables $\mathbf{S}$, where $S_j$ indicates whether node $j$ is a parent in the structure of node $i$. 

Structure learners then calculate $\mathcal{B}$ frequently during execution time to score graphs. Alternatively, they are also used to score topological orderings, as many structure learning methods, such as TRUST \cite{wang2022trust} and ArCO-GP \cite{toth2025effectivebayesiancausalinference}, consider the smaller space of topological orderings. Given an ordering $\sigma = (\sigma_1, \sigma_2, \dotsc, \sigma_n)$ of the nodes $X_1, X_2, \dotsc, X_n$, the unnormalized posterior probability of the ordering is given by summing the scores of all graphs consistent with that ordering:
\[
\sum_{i=1}^n \sum_{\mathbf{S} \,:\, S_{\sigma_j} = 0 \ \forall j \geq i} \mathcal{B}_{\sigma_i}(\mathbf{S})
\]
This summation enforces the restriction of topological ordering: $\sigma_j$ cannot be a parent of $\sigma_i$ if $j \geq i$. 

This equation highlights why marginalization is a critical task in structure learning, since we calculate the inner summation with a marginal query on the function $\mathcal{B}_{\sigma_i}$. We query for the probability mass where the variables $S_{\sigma_j}$ for $j < i$ are marginalized and for $j \geq i$, $S_{\sigma_j} = 0 $. As learners may score many orderings during their execution, we require efficient marginalization over \emph{any} set of variables. We define these types of queries as \textit{marginal/zero} queries, as all possible $S_j$ are either set to $0$ or are marginalized variables. These types of queries thus play an especially important role in our proposed learning procedure.

\section{Methods}

Throughout this section, we describe and examine the application of our marginalization method on a single variable $X_i$. Thus, in the rest of this section, we omit the subscript $X_i$ from $\mathcal{B}_{X_i}$ because we assume that we are marginalizing the score for $X_i$. In practice, however, we repeat our method for every variable in the data. 

\subsection{Exact Computation}

\citet{viinikka2020scalablebayesianlearning} propose an algorithm that computes using dynamic programming all possible marginal probabilities in $O(3^N)$ time. This algorithm precomputes all possible marginal masses, stores the results in a lookup table, and then retrieves these results in $O(1)$ time to answer queries. We discuss the methodology of these methods, as they are used to compute parts of the ground-truth data for our learning dataset. 

For disjoint subsets $A_i, A_i' \subseteq \{1, 2, \dotsc, n\}$ and $A_i \cap A_i' = \emptyset$, we define \[g(A_i, A_i') = \sum_{\mathbf{S} \, : \, \forall i \in A_i, S_i = 1, \forall i \in A_i', S_i' = 0} \mathcal{B}(\mathbf{S})\]

In other words, this corresponds to the probability mass where all variables whose indices in $A_i$ must be $1$, all variables whose indices in $A_i'$ must be $0$, and all other variables are marginalized over. This definition lends itself to the handy recurrence relation \[g(A_i, A_i') = g(A_i \cup b, A_i') + g(A_i, A_i' \cup b)\]
for any $b \in \{1, 2, \dotsc, n\} \setminus(A_i \cup A_i')$. This relation holds true because when we marginalize over a variable, there are still two disjoint cases: that variable has value $1$ and that variable has value $0$.

As for base cases, if $A_i \cup A_i' = \{1, 2, \dotsc, n \}$, then let the vector $\mathbf{K}$ be the vector where all indices in $A_i$ are $1$ and where all indices in $A_i'$ are $0$. There are no marginalized variables in this case, so $g$ is fully specified. Since we have no variables to marginalize over, we can simply set  
$g(A_i, A_i') = \mathcal{B}(\mathbf{K})$.

We can thus combine the above base cases with the recurrence relation to compute all marginal probability masses. There are $3^N$ states, since each variable may either be set to $0$, set to $1$, or marginalized over, so this algorithm runs in $O(3^N)$ time and requires $O(3^N)$ memory to store the table. Thus, when $N$ is greater than around 16, these methods become computationally infeasible, so existing methods limit to considering only a subset of candidate parents $\mathbf{S_C} \subseteq \mathbf{S}$. These candidate parents are chosen to gain maximum coverage over the parent set distribution; a variety of potential heuristics to do so are evaluated in \citet{viinikka2020scalablebayesianlearning}. Inevitably, some possible parent sets are ruled out by this approach.

\subsection{Probabilistic Circuit}
In order to gain support over all possible parent sets, we would like to learn some sort of probabilistic model or neural network that could approximate $\mathcal{B}$. However, for most architectures, we would then no longer be able to marginalize in sub-exponential time. Thus, we instead propose to use a decomposable and smooth PC as our architecture. We can then answer marginal queries in a time linear to the size of the circuit, while also avoiding incurring the exponential $O(3^N)$ precomputation cost. 

Our circuit is designed to output unnormalized probabilities, since the scores that we want to learn for structure learning are unnormalized. PCs are often  designed to enforce normalization by ensuring $\sum w_i = 1$ for each sum node (e.g. by pushing the parameterization through a softmax). We simply remove this restriction within our sum nodes.\footnote{We do store our parameters in log-domain, as is the norm in PC learning for reasons of numerical stability and ensuring non-negativity.} However, computing the distribution of an unnormalized PC is still easy: we can tractably marginalize over all variables to find our normalizing constant.

\subsubsection{PC Structure}
The structure of our PC is similar to that of RAT-SPN, proposed in \citet{peharz2020ratspn}. We parameterize our PC by a latent size $N$ along with a number of variables $M$ (in our method, $M$ will always be one less than the total number of variables in the BN $\mathcal{N}$ since we do not consider the node we are currently scoring). We begin with a leaf layer of nodes at the bottom, and then alternate between product and sum layers. For our PC, each layer is organized as a matrix of nodes, with this matrix always having $N$ columns of nodes. When computing across our entire PC, we maintain the invariant that all rows of nodes have the same scope while columns of nodes have pairwise disjoint scopes. The structure of the PC is then as following:

\begin{algorithm}[t]
\hspace*{\algorithmicindent} \textbf{Input:} Bayesian scorer $\mathcal{B}$, candidate parent set $C$, marginal training limit $L$
\caption{PC learning}
\label{alg:cap}
\begin{algorithmic}[1]
\STATE initialize a probabilistic circuit $p$
\STATE $S_{b,\text{train}}, S_{b,\text{val}} \gets$ randomly sampled parent sets from $\mathbf{S}$
\STATE score $S_{b,\text{train}}, S_{b,\text{val}}$ with $\mathcal{B}$
\FOR {number of baseline epochs}
\STATE {train $p$ with gradient descent on $S_{b,\text{train}}$ using MSE loss}
\ENDFOR
\STATE {train an exact DP marginalizer $M$ on the limited parent set $C$}
\FOR {$i = 1$ to $L$}
\STATE $S_{m,\text{train}}, S_{m,\text{val}} \gets$ randomly sampled $(i, 0)$ and $(i, 1)$ marginals from $\mathbf{S}$
\STATE use $M$ to generate labels for $S_{m,\text{train}}, S_{m,\text{val}}$
\STATE $S_{n,\text{train}}, S_{n,\text{val}} \gets$ randomly sampled parent sets from $\mathbf{S}$
\STATE score $S_{n,\text{train}}, S_{n,\text{val}}$ with $\mathcal{B}$
\STATE merge $S_{m,\text{train}}$ with $S_{n,\text{train}}$ and $S_{m,\text{val}}$ with $S_{n,\text{val}}$
\FOR{ number of finetune epochs}
\STATE {train $p$ with gradient descent on combined data set using MSE loss}
\ENDFOR
\ENDFOR
 \end{algorithmic}
\end{algorithm}

\begin{enumerate}
\item  The leaf layer forms a matrix of leaf nodes of shape $(M, N)$. Each leaf node's value corresponds to the presence or absence of a certain parent within the parent set. The leaf nodes of our PC take parameterized, unnormalized Bernoulli random variables as input distributions. Within our leaf layer, we create $N$ repeated leaf nodes for each of our $M$ variables. The order of the $M$ variables is determined by a random permutation, which is fixed across all $N$ repetitions. This design ensures that the variable scopes are combined in a consistent yet randomized manner within product nodes, while keeping the scope of sum nodes consistent.

\item The input to a product layer is the output of either a leaf layer or a sum layer. The product layer doubles the amount of variables in the scope of each node and thus cuts the number of rows of our matrix in half. Given that the previous layer was a grid of $S = (A, N)$ nodes, our product layer has dimensionality $(A/2, N)$. Within each product node, $P_{i,j} = S_{2i,j} * S_{2i+1, j}$. This enforces the constraint of decomposability given that columns of nodes have pairwise disjoint scopes.

\item The sum layer creates a weighted sum of input nodes from the same row without affecting the scope of nodes. Given that the previous layer was a grid of $P = (A, N)$ nodes, our sum layer has dimensionality $(A, N)$. Within each sum node, $S_{i,j} = \sum_{k = 1}^N w_{i, j, k} * P_{i, k}$. This construction enforces the constraint of smoothness, given that rows of nodes have the same scope. The tensor $w$ of dimension $(A, N, N)$ contains parameters that we learn during training. 
\end{enumerate}

\subsection{PC Learning}
Remember that we define a marginal/zero query as one where we marginalize over some variables and require that the rest be $0$. We further define a $(k, 0)$ query as a marginal/zero query that marginalizes over exactly $k$ variables and sets the rest to $0$ and a $(k, 1)$ query as a query that marginalizes over $k$ variables, sets one variable to $1$, and sets the rest to $0$. 

Given these definitions, we present a two-phase learning routine in algorithm $\ref{alg:cap}$. We first begin with a baseline learning phase that trains on complete parent sets whose scores have been calculated by the Bayesian scorer $\mathcal{B}$. 

Then, during the second phase, we finetune the model on a combination of $(k, 0)$ and $(k, 1)$ marginals along with complete unmarginalized parent sets. During this phase, we increase $k$ iteratively. We primarily train on marginal/zero queries as they are one of the principal types of queries that structure learning methods require us to answer. We also train in increasing number of marginalized variables, as successfully training each previous iteration helps training with the next iteration. According to the recurrence relation written in the exact computation section, each $(k+1, 0)$ query can be written as a sum of a $(k, 0)$ query and a $(k, 1)$ query, as the additional marginalized variable must either be set to $0$ and $1$ when summing. Accordingly, training successfully at a certain $k$ can intuitively help to successfully build our marginal queries at $k+1$. 

During both learning phases, we used the mean squared error (in log-domain) as our loss function. That is:
$$
(\log p(\mathbf{S})  - \mathcal{B}(\mathbf{S}))^2
$$
MSE has also been used in similar models focused on learning marginals \cite{liu2024generativemarginalizationmodels} and was the best choice empirically compared to other loss choices like KL divergence.

\begin{figure*}[t]
\centering
\includegraphics[width=0.9\textwidth]{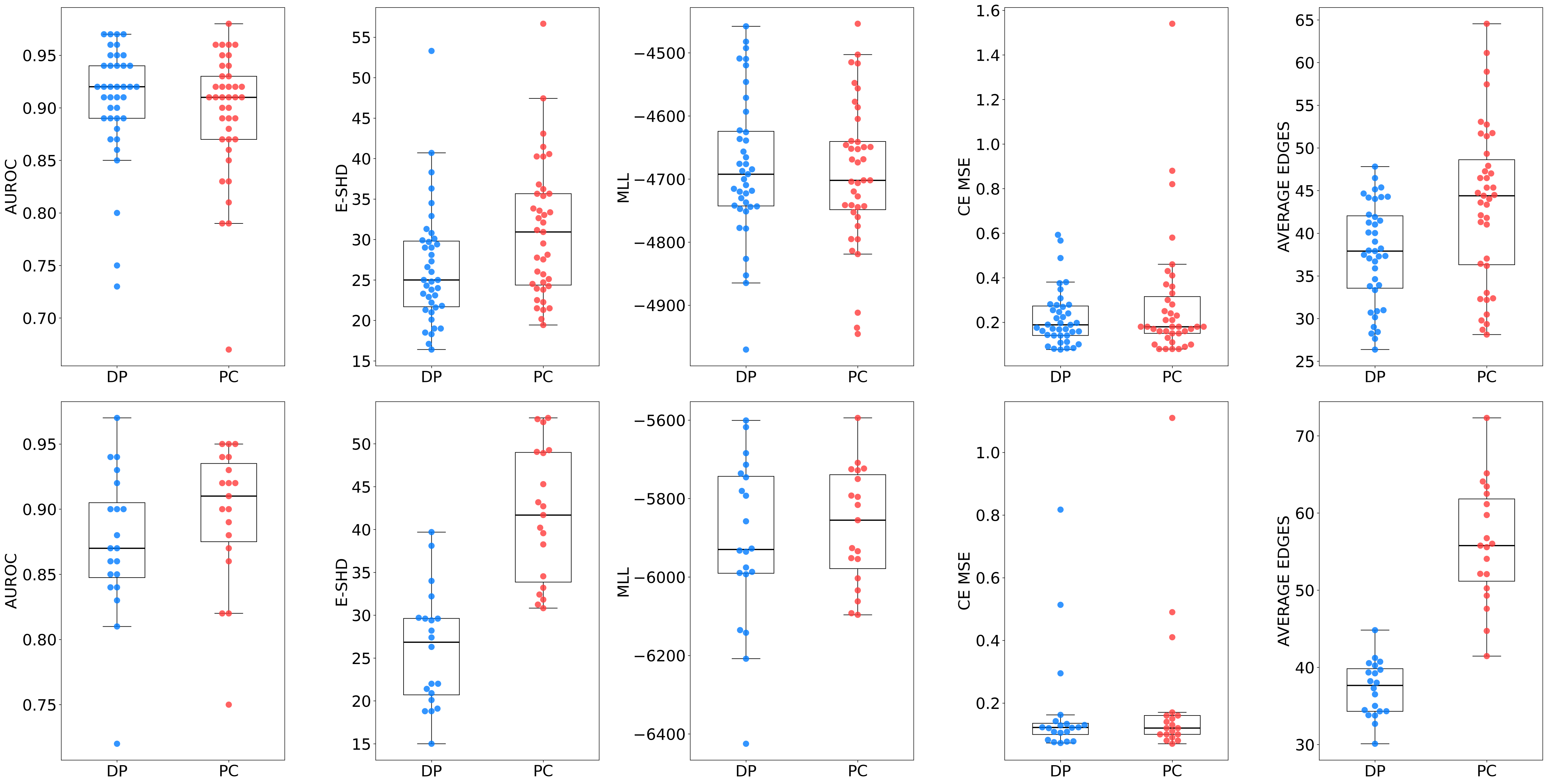}
\caption{ Evaluation running TRUST between the exact DP marginalization and regression circuits \textbf{Top:} Comparison on $d = 16$. The DP marginalization here considers all 15 other nodes, so marginals are calculated exactly. Our approximate regression circuit performs comparably to the DP, except for E-SHD. \textbf{Bottom:} Comparison on $d = 20$. We now restrict the DP to considering only $8$ nodes, as is normal practice. As expected, our PC performs considerably better now on all metrics except E-SHD.}
\label{fig1}
\end{figure*}

\subsection{Additional Implementation Details}
We also make the following empirical choices for our method.
\begin{enumerate}
    \item For the first phase, we do not sample parent set vectors uniformly. Instead, we sample parent set vectors from $\mathbf{S}$ where each vector is weighted by how many zero/marginal probability masses they contribute to. A parent vector with $T$ indicators equal to $1$ will contribute to $2^{M-T}$ marginal probability masses (since all other indicators must either be marginalized over or set equal to $0$). Thus, its weight in the sampling is $\propto  2^{M-T}$. 
    \item We combine both marginals and complete parent sets in the second learning phase because only learning marginals may cause our regression circuit to deviate from already learned complete parent set values. Moreover, since we resample our complete parent sets in this step, the regression circuit can train on additional previously unseen training data.
    \item Learning rate is set relatively high at $10^{-1}$ for the first phase. This helps alleviate a problem with vanishing gradients that is exacerbated by the fact that our weights are unnormalized and so can take on increasingly negative values in the log domain. We reduce this learning rate upon training loss plateau to avoid unstable convergence.
    \item We also have a limitation of $L$ for our second phase iteration for the purposes of computation efficiency. Moreover, having too many iterations causes the regression circuit to forget the marginals it has learned in previous iterations, leading to decreased effectiveness. 
    \item Correct initialization of the parameters of the regression circuit is important for gradient-based optimization. With a multiplier hyperparameter $m < 0$, we initialize our parameters (in log-domain) according to the distribution $m\log(U(0, 1))$. In particular we set $m$ to be around $-10$, as this setting encourages the breaking of gradient symmetry during training.
\end{enumerate}
\section{Experiments}

We perform an empirical evaluation of our method within the context of the TRUST framework \cite{wang2022trust}. TRUST is a Bayesian structure learning method that relies on probabilistic circuits. The original TRUST still employs dynamic programming marginalization within its leaf nodes. Thus, we replace these DP calculations with our method. We chose to work with TRUST as it also utilizes probabilistic circuits as a framework for structure learning, which allows us to create a method that utilizes entirely probabilistic circuits. 

We maintain similar experimental settings as \citet{wang2022trust}. Namely, we utilize synthetically generated data, where the ground truth structures are Erdős–Rényi graphs of size $d \in \{16, 20\}$ with an average of $2$ edges per node. The constructed Bayesian networks use linear Gaussian mechanisms. Accordingly, we utilize the BGe score as our Bayesian score. We generate $100$ samples for our training data $D_{\text{train}}$.

\subsection{Evaluation}

We evaluated our structure learning experiments with the following metrics, which are the same as presented in \citet{wang2022trust}. Let the ground truth graph be $G$ with our posterior distribution represented by $q$. We hold $1000$ samples for our test data $D_{\text{test}}$.

The \textit{expected structural Hamming distance} (E-SHD) measures the expected number of different edges between the completed partially directed acyclic graphs (CPDAGs) of the ground truth $G$ and of graphs $G'$ sampled from our posterior $q$. CPDAGs are the structure that describe the Markov equivalence class of a DAG. Thus, we should compare the CPDAGs of $G$ and $G'$ against each other, not just the original DAGs.

The \textit{area under the receiver operating curve} (AUROC) for structure learning is evaluated by first computing the marginal probabilities within our posterior distribution of every potential edge in the graph. Then, we vary a confidence threshold to construct a ROC curve \cite{friedman2003beingbayesianaboutnetwork}.

The \textit{marginal log likelihood} (MLL) metric utilizes the BGE score to measure how likely our held out test data $D_{\text{test}}$ is under our constructed posterior distribution $q$.

The \textit{mean squared error of causal effects} (MSE-CE) is computed as the average mean squared error between the true and predicted causal effects across all pairs of nodes.

In Figure \ref{fig1}, we compare on $d = 16$ the results of TRUST with our PC method against a DP method with no candidate set restrictions. We ran these experiments between DP and PC methods with the same DAGs and data, allowing for direct comparison. Our baseline circuit had latent size $N = 256$ and initialization multiplier $m = 12$. For both the first and second steps of training, we utilized a batch size of $500$. In the first learning phase of training, we used $10000$ samples for training and $1000$ samples for validation, under a learning rate of $10^{-1}$. Within the second step of training, we used a training dataset of $20000$ samples, evenly divided between marginals and complete scores, with a validation dataset of $2000$, also split evenly. For this phase, we set the learning rate to $5*10^{-3}$, used $L = 7$ iterations, and trained with $20$ epochs per iteration.

Since we did not restrict parent sets for the DP for this experiment, the DP will always calculate exact marginals. In other words, the size of candidate sets $|\mathbf{S_c}| = 15$, so this baseline serves as the posterior that we intend to approximate. Although our learning procedure for our circuit only learns the distribution of scores approximately, our method achieves performance comparable to the exact calculation, except for E-SHD. This result shows that we can approximate the exact marginals well enough to have limited effect on downstream TRUST performance.

Once we restrict the candidate set for the DP method, as is standard practice, our method showcases improved performance. On $d = 20$ while restricting the candidate parent set to $8$ nodes, our method performs better on all metrics except for E-SHD. In this case, compared to the PC settings on $d = 16$, we reduce the number of latents of our PC from $256$ to $64$, increase our batch size to $1000$ samples, and double the size of all training and validation sets for the PC. We keep the rest of the settings the same.

As for why our method holds a higher E-SHD than the exact computation in both $d = \{16, 20\}$, our method tends to sample graphs with more edges, as seen in the rightmost plots in Figure \ref{fig1}. However, the AUROC scores indicate that the PC-based posterior is nonetheless capable of determining which edges are more likely. 
\section{Conclusion}

We show that probabilistic circuits can be used to perform marginalization tasks in Bayesian structure learning. To do so, we learn a distribution for the circuit that approximates a Bayesian score. Once this distribution has been learned, we can exactly marginalize over it. Compared to the dynamic programming approach which exactly marginalizes over a limited subset of the ground truth distribution, our method allows us to have support over all parent sets. We show that this approach yields performance improvements by creating a more accurate posterior distribution.

As for future avenues of work, one important focus is to scale to higher dimensions for TRUST, as our method removes the restriction on the number of candidate parents that previously prevented such scaling. Additionally, limited testing has shown that this restriction hampers performance more significantly when trying to learn denser structures (such as an Erdős–Rényi graph with an average of $4$ edges per node). Experimenting with denser graphs may thus show that utilizing our method yields larger performance improvements than on sparser graphs.

Moreover, our marginalization circuit is limited in its applicability to just TRUST or linear Gaussian BNs. As previously mentioned, ArCO-GP \cite{toth2025effectivebayesiancausalinference} and other methods also rely on limits for the number of candidate parents of a node. As such, an important avenue for future work is to adapt our approach to using different BN mechanisms and thus scores in order to be compatible with these other structure learning algorithms.

Overall, we have presented a promising new approach to the computational problem of marginalizing Bayesian scores, that \emph{learns} a tractable approximation to the ground truth score function that then enables efficient answering of marginal queries on-the-fly. We show that this naturally outperforms existing methods, which impose hard, hand-crafted limits for computational feasibility, such as restricting the candidate parent sets.
\bibliography{ref.bib}
\end{document}